\documentclass[11pt]{article}
\usepackage{authblk}
\usepackage{eacl2017}
\usepackage{times}
\usepackage{url}
\usepackage{latexsym}
\usepackage{bm}
\usepackage{graphicx,color,xcolor}
\usepackage{latexsym}
\usepackage{amsmath}
\usepackage{pifont}
\usepackage{amsfonts}
\usepackage{dsfont}
\usepackage{multirow}
\usepackage{booktabs, caption}
\usepackage{tabularx}
\usepackage{listliketab}
\usepackage{lipsum} 

\usepackage{mathrsfs}
\usepackage{algorithm}
\usepackage{enumitem}
\usepackage{subfigure}
\usepackage{amssymb}
\usepackage[noend]{algpseudocode}

\usepackage{xcolor}

\DeclareMathOperator*{\argmax}{arg\,max}

\makeatletter
\def\BState{\State\hskip-\ALG@thistlm}
\makeatother
\eaclfinalcopy 
\def\eaclpaperid{199} 

\makeatletter
\renewcommand\AB@affilsepx{ \protect\Affilfont}
\makeatother

\title{Learning to Translate in Real-time with Neural Machine Translation}
\def \nyu{$^\ddag$}
\def \hku{$^\dagger$}
\def \cmu{$^\lozenge$}
\author[\hku]{\bf Jiatao Gu}
\author[\cmu]{\bf Graham Neubig}
\author[\nyu]{\bf Kyunghyun Cho}
\author[\hku]{\bf Victor O.K. Li}
\affil[\hku]{The University of Hong Kong}
\affil[\cmu]{Carnegie Mellon University}
\affil[\nyu]{New York University\authorcr }
\affil[\hku]{\tt  \{jiataogu, vli\}@eee.hku.hk}
\affil[\cmu]{\tt  gneubig@cs.cmu.edu\authorcr }
\affil[\nyu]{\tt  kyunghyun.cho@nyu.edu}

\date{}

\begin{document}
\maketitle
\begin{abstract}
Translating in real-time, a.k.a.~simultaneous translation, outputs translation words before the input sentence ends,
which is a challenging problem for conventional machine translation methods. 
We propose a neural machine translation (NMT) framework for simultaneous translation in which an agent learns to make decisions on when to translate from the interaction with a pre-trained NMT environment.
To trade off quality and delay, we extensively explore various targets for delay and design a method for beam-search applicable in the simultaneous MT setting. Experiments against state-of-the-art baselines on two language pairs demonstrate the efficacy of the proposed framework both quantitatively and qualitatively.%
\footnote{Code and data can be found at \url{https://github.com/nyu-dl/dl4mt-simul-trans}.}



\end{abstract}

\section{Introduction}
Simultaneous translation, the task of translating content in real-time as it is produced, is an important tool for real-time understanding of spoken lectures or conversations \cite{fugen2007simultaneous,bangalore2012real}.
Different from the typical machine translation~(MT) task, in which translation quality is paramount, simultaneous translation requires balancing the trade-off between translation quality and time delay to ensure that users receive translated content in an expeditious manner \cite{mieno2015speed}.
A number of methods have been proposed to solve this problem, mostly in the context of phrase-based machine translation.
These methods are based on a \textit{segmenter}, which receives the input one word at a time, then decides when to send it to a MT system that translates each segment independently \cite{oda-EtAl:2014:P14-2} or with a minimal amount of language model context \cite{bangalore2012real}.
\begin{figure}[!t]
   	\centering
          	\includegraphics[width=\linewidth]{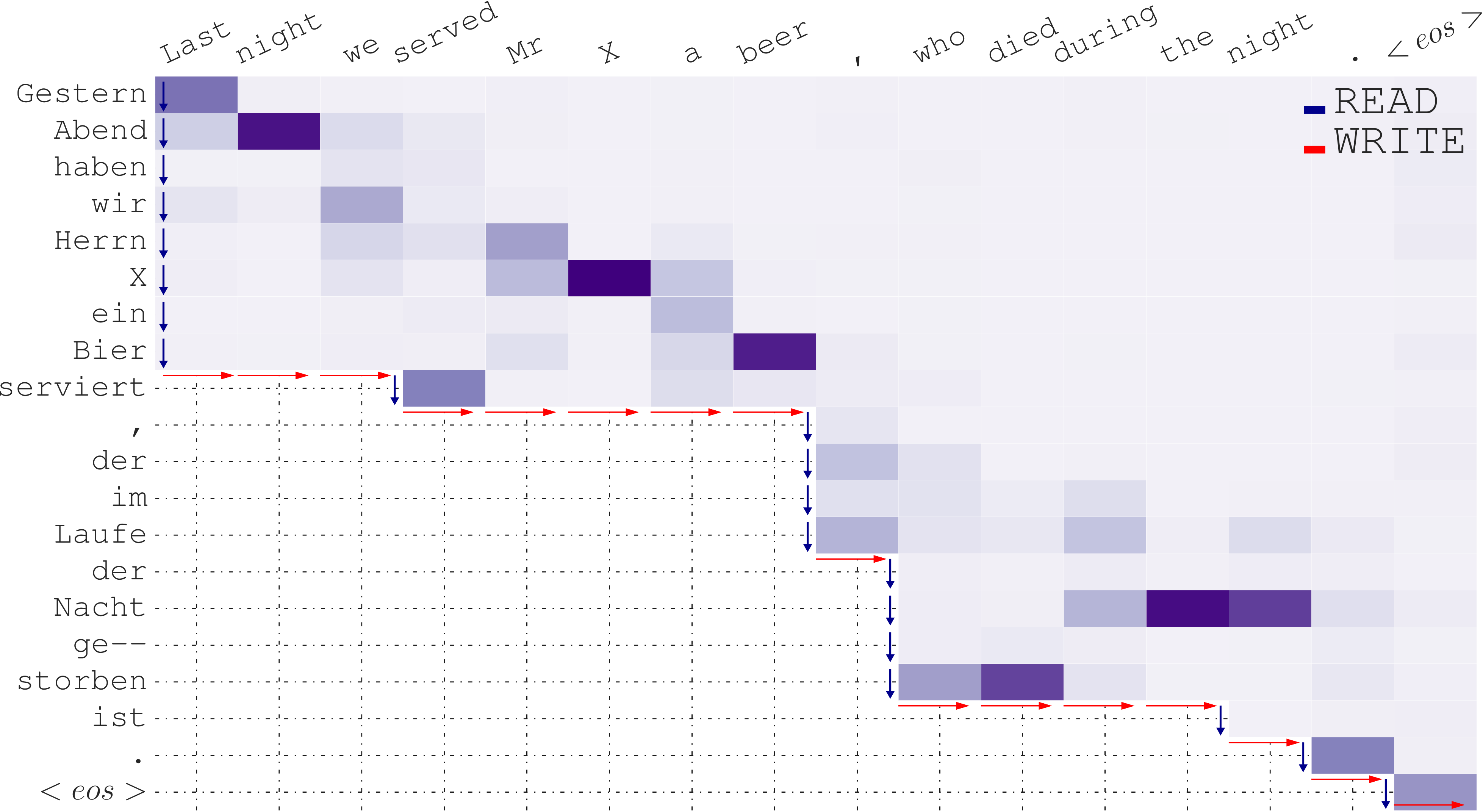} 
            \vspace{-15pt}
          	\caption{\label{crop} {Example output from the proposed framework in DE $\rightarrow$ EN simultaneous translation. The heat-map represents the soft alignment between the incoming source sentence (left, up-to-down) and the emitted translation (top, left-to-right). The length of each column represents the number of source words being waited for before emitting the translation. Best viewed when zoomed digitally.}}
\vspace{-6mm}
\end{figure} 

Independently of simultaneous translation, accuracy of standard MT systems has greatly improved with the introduction of neural-network-based MT systems (NMT)~\cite{sutskever2014sequence,bahdanau2014neural}.
Very recently, there have been a few efforts to apply NMT to simultaneous translation either through heuristic modifications to the decoding process \cite{cho2016can}, or through the training of an independent segmentation network that chooses when to perform output using a standard NMT model \cite{satija2016simultaneous}.
However, the former model lacks a capability to learn the appropriate timing with which to perform translation, and the latter model uses a standard NMT model as-is, lacking a holistic design of the modeling and learning within the simultaneous MT context.
In addition, neither model has demonstrated gains over previous segmentation-based baselines, leaving questions of their relative merit unresolved.

In this paper, we propose a unified design for learning to perform neural simultaneous machine translation.
The proposed framework is based on formulating translation as an interleaved sequence of two actions: \textsc{read} and \textsc{write}.
Based on this, we devise a model connecting the NMT system and these \textsc{read}/\textsc{write} decisions.
An example of how translation is performed in this framework is shown in Fig.~\ref{crop}, and detailed definitions of the problem and proposed framework are described in \S\ref{sec:definition} and \S\ref{sec:framework}.
To learn which actions to take when, we propose a reinforcement-learning-based strategy with a reward function that considers both quality and delay (\S\ref{sec:optimization}).
We also develop a beam-search method that performs search within the translation segments (\S\ref{sec:beamsearch}).

We evaluate the proposed method on English-Russian~(EN-RU) and English-German~(EN-DE) translation in both directions (\S\ref{sec:experiments}).
The quantitative results show strong improvements compared to both the NMT-based algorithm and a conventional segmentation methods.
We also extensively analyze the effectiveness of the learning algorithm and the influence of the trade-off in the optimization criterion, by varying a target delay.
Finally, qualitative visualization is utilized to discuss the potential and limitations of the framework.

\section{Problem Definition}
\label{sec:definition}
\vspace{-5pt}

Suppose we have a buffer of input words $X = \{x_1, ..., x_{T_s}\}$ to be translated in real-time. We define the simultaneous translation task as sequentially making two interleaved decisions: \textsc{read} or \textsc{write}.
More precisely, the translator \textsc{read}s a source word $x_\eta$ from the input buffer in chronological order as translation context, or \textsc{write}s a translated word $y_\tau$ onto the output buffer, resulting in output sentence $Y = \{y_1, ..., y_{T_t}\}$,
and action sequence $A=\{a_1, ..., a_T\}$ consists of $T_s$ \textsc{read}s and $T_t$ \textsc{write}s, so $T=T_s+T_t$.

Similar to standard MT, we have a measure $Q(Y)$ to evaluate the translation quality, such as BLEU score~\cite{papineni2002bleu}.
For simultaneous translation we are also concerned with the fact that each action incurs a time delay $D(A)$.
$D(A)$ will mainly be influenced by delay caused by \textsc{read}, as this entails waiting for a human speaker to continue speaking (about $0.3$s per word for an average speaker), while \textsc{write} consists of generating a few words from a machine translation system, which is possible on the order of milliseconds.
Thus, our objective is finding an optimal policy that generates decision sequences with a good trade-off between higher quality $Q(Y)$ and lower delay $D(A)$.
We elaborate on exactly how to define this trade-off in \S\ref{sec:reward}.



In the following sections, we first describe how to connect the \textsc{read}/\textsc{write} actions with the NMT system (\S\ref{sec:framework}), and how to optimize the system to improve simultaneous MT results (\S\ref{sec:optimization}).

\section{Simultaneous Translation \\ with Neural Machine Translation}
\label{sec:framework}

\begin{figure}[!t]
   	\centering
          	\vspace{-10pt}
          	\includegraphics[width=\linewidth]{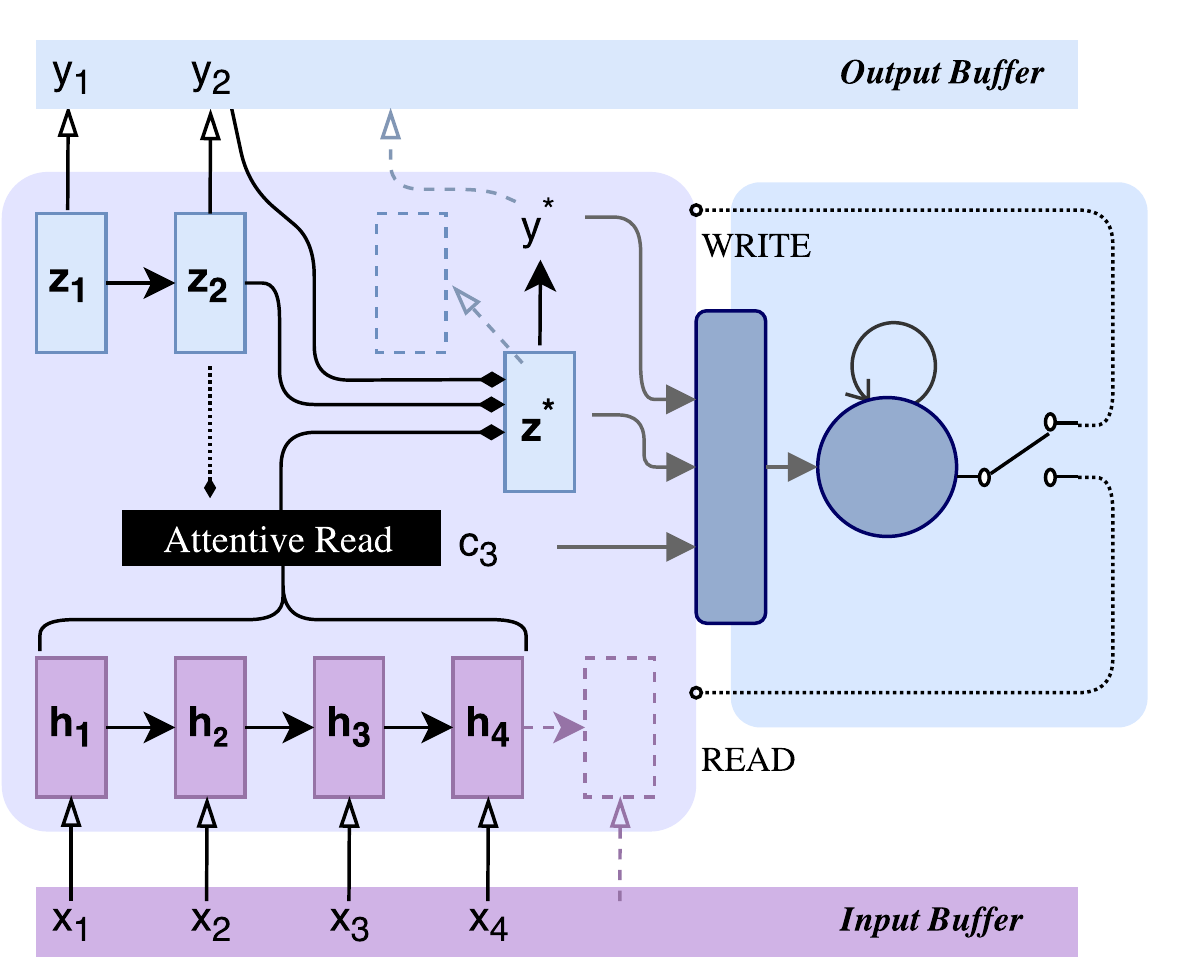} 
          	\vspace{-15pt}
          	\caption{\label{snmt} {Illustration of the proposed framework: at each step, the NMT environment (left) computes a candidate translation. The recurrent agent (right) will the observation including the candidates and send back decisions--\textsc{read} or \textsc{write}.}} 
   \vspace{-12pt}
  \end{figure}

The proposed framework is shown in Fig.~\ref{snmt}, and can be naturally decomposed into two parts: environment (\S\ref{sec:environment}) and agent (\S\ref{sec:agent}).
  
\subsection{Environment}
\label{sec:environment}
\paragraph{Encoder:~\textsc{read}} 
The first element of the NMT system is the encoder, which converts input words $X=\{x_1, ..., x_{T_s}\}$ into context vectors $H = \{h_1, ..., h_{T_s}\}$.
Standard NMT uses bi-directional RNNs as encoders \cite{bahdanau2014neural}, but this is not suitable for simultaneous processing as using a reverse-order encoder requires knowing the final word of the sentence before beginning processing.
Thus, we utilize a simple left-to-right uni-directional RNN as our encoder:
\vspace{-5pt}
\begin{equation}
\label{enc}
    h_{\eta} = \phi_{\textsc{uni-enc}}\left(h_{\eta-1}, x_{\eta}\right)
\vspace{-4pt}
\end{equation}

\paragraph{Decoder:~\textsc{write}} Similar with standard MT, we use an attention-based decoder. In contrast, we only reference the words that have been read from the input when generating each target word:
\vspace{-5pt}
\begin{equation}
\label{dec}
    \begin{split}
        &c_{\tau}^{\eta} = \phi_{\textsc{att}}\left(z_{\tau-1}, y_{\tau-1}, H^{\eta}\right) \\
        &z_{\tau}^{\eta} = \phi_{\textsc{dec}}\left(z_{\tau-1}, y_{\tau-1}, c_{\tau}^{\eta}\right)\\
        &p\left(y|y_{<\tau}, H^{\eta}\right) \propto \exp\left[\phi_{\textsc{out}}\left(z_{\tau}^{\eta}\right)\right],
    \end{split}
\vspace{-4pt}
\end{equation}
where for $\tau$, $z_{\tau-1}$ and $y_{\tau-1}$ represent the previous decoder state and output word, respectively.
$H^{\eta}$ is used to represent the incomplete input states, where $H^{\eta}$ is a prefix of $H$.  
As the \textsc{write} action calculates the probability of the next word on the fly, we need greedy decoding for each step:
\vspace{-5pt}
\begin{equation}
    y_{\tau}^{\eta} = \argmax\nolimits_{y}p\left(y|y_{<\tau}, H^{\eta}\right) 
\vspace{-2pt}
\end{equation}
Note that $y_{\tau}^{\eta}, z_{\tau}^{\eta}$ corresponds to $H^{\eta}$ and is the candidate for $y_{\tau}, z_{\tau}$.
The agent described in the next section decides whether to take this candidate or wait for better predictions.

\subsection{Agent}
\label{sec:agent}
A trainable agent is designed to make decisions $A=\{a_1, .., a_T\}, a_t \in \mathcal{A}$ sequentially based on observations $O=\{o_1, ..., o_T\}, o_t \in \mathcal{O}$, and then control the translation environment properly. 
\vspace{-3pt}
\paragraph{Observation} As shown in Fig~\ref{snmt}, we concatenate the current context vector $c_{\tau}^{\eta}$, the current decoder state $z_{\tau}^{\eta}$ and the embedding vector of the candidate word $y_{\tau}^{\eta}$ as the continuous observation, $o_{\tau+\eta}=\left[c_{\tau}^{\eta}; z_{\tau}^{\eta}; E(y_{\tau}^{\eta})\right]$ to represent the current state.

\paragraph{Action} Similarly to prior work~\cite{grissomii2014don}, we define the following set of actions:
\begin{itemize}[leftmargin=*]
\vspace{-2mm}
	\itemsep -0.4em
    \item \textbf{\textsc{read}}: the agent rejects the candidate and waits to encode the next word from input buffer;
    \item \textbf{\textsc{write}}: the agent accepts the candidate and emits it as the prediction into output buffer;
\end{itemize}

\paragraph{Policy} How the agent chooses the actions based on the observation defines the policy. In our setting, we utilize a stochastic policy $\pi_{\theta}$ parameterized by a recurrent neural network, that is:
\vspace{-5pt}
\begin{equation}
    \begin{split}
        &s_t = f_{\theta}\left(s_{t-1}, o_t\right)\\
        &\pi_{\theta}(a_t|a_{<t}, o_{\leq t}) \propto g_{\theta}\left(s_t\right), 
    \end{split}
    \vspace{-2pt}
\end{equation}
where $s_t$ is the internal state of the agent, and is updated recurrently yielding the distribution of the action $a_t$. 
Based on the policy of our agent, the overall algorithm of greedy decoding is shown in Algorithm~\ref{algo1}, 
The algorithm outputs the translation result and a sequence of observation-action pairs.

\begin{algorithm}
\caption{Simultaneous Greedy Decoding}
\label{algo1}
\begin{algorithmic}[1]
\Require{NMT system $\phi$, policy $\pi_{\theta}$, $\tau_{\textsc{max}}$, input buffer $X$, output buffer $Y$, state buffer $S$.}
\State \textbf{Init} $x_1 \Leftarrow X, h_1 \gets \phi_{\textsc{enc}}\left(x_1\right), H^1 \gets \{h_1\}$
\State \hspace{17pt} $z_0 \gets \phi_{\textsc{init}}\left(H^1\right), y_0 \gets \left<s\right>$
\State \hspace{17pt} $\tau \gets 0, \eta \gets 1$
\While{$\tau < {\tau}_{\textsc{max}}$}
\State $t \gets \tau + \eta$
\State $y_{\tau}^{\eta}, z_{\tau}^{\eta}, o_t \gets \phi\left(z_{\tau-1}, y_{\tau-1}, H^{\eta}\right)$ 
\State $a_t \sim \pi_{\theta}\left(a_t; a_{<t}, o_{<t}\right), S \Leftarrow (o_t, a_t)$
\If {$a_t=\textsc{read}$ and $x_{\eta} \neq \left</s\right>$}
\State $x_{\eta+1} \Leftarrow X, h_{\eta+1} \gets \phi_{\textsc{enc}}\left(h_{\eta}, x_{\eta+1}\right)$
\State $H^{\eta+1} \gets H^{\eta} \cup \{h_{\eta+1} \}, \eta \gets \eta + 1$
\If {$|Y|=0$} $z_0 \gets \phi_{\textsc{init}}\left(H^{\eta}\right)$
\EndIf
\ElsIf {$a_t = \textsc{write}$}
\State $z_{\tau} \gets z_{\tau}^{\eta}, y_{\tau} \gets y_{\tau}^{\eta}$ 
\State $Y \Leftarrow y_{\tau}, \tau \gets \tau + 1$
\If {$y_{\tau}=\left</s\right>$} \textbf{break}
\EndIf
\EndIf
\EndWhile
\end{algorithmic}
\end{algorithm}

\section{Learning}
\label{sec:optimization}
The proposed framework can be trained using reinforcement learning. More precisely, we use policy gradient algorithm together with variance reduction and regularization techniques.
\subsection{Pre-training} 
We need an NMT environment for the agent to explore and use to generate translations.
Here, we simply pre-train the NMT encoder-decoder on full sentence pairs with maximum likelihood, and assume the pre-trained model is still able to generate reasonable translations even on incomplete source sentences.
Although this is likely sub-optimal, our NMT environment based on uni-directional RNNs can treat incomplete source sentences in a manner similar to shorter source sentences and has the potential to translate them more-or-less correctly.



\vspace{-3pt}
\subsection{Reward Function}
\label{sec:reward}

The policy is learned in order to increase a reward for the translation. At each step the agent will receive a reward signal $r_t$ based on $(o_t, a_t)$. To evaluate a good simultaneous machine translation, a reward must consider both quality and delay. 
\vspace{-3pt}
\paragraph{Quality} We evaluate the translation quality using metrics such as BLEU~\cite{papineni2002bleu}. The BLEU score is defined as the weighted geometric average of the modified n-gram precision $\textsc{BLEU}^0$, multiplied by the brevity penalty $\textsc{BP}$ to punish a short translation. In practice, the vanilla BLEU score is not a good metric at sentence level because being a geometric average, the score will reduce to zero if one of the precisions is zero. To avoid this, we used a smoothed version of BLEU for our implementation~\cite{lin2004automatic}.
\vspace{-3pt}
\begin{equation}
    \textsc{BLEU}(Y, Y^*) = \textsc{BP} \cdot \textsc{BLEU}^0(Y, Y^*),
    \vspace{-3pt}
\end{equation}
where $Y^*$ is the reference and $Y$ is the output.  
We decompose $\textsc{BLEU}$ and use the difference of partial $\textsc{BLEU}$ scores as the reward, that is:
\begin{equation}
    r_t^Q=\left\{
\begin{array}{rcl}
\Delta\textsc{BLEU}^0(Y, Y^*, t)     &      & {t < T}\\
\textsc{BLEU}(Y, Y^*)    &      & {t = T}
\end{array} \right. 
\end{equation}
where $Y^t$ is the cumulative output at $t$ ($Y^0=\emptyset$), and $\Delta\textsc{BLEU}^0(Y, Y^*, t) = \textsc{BLEU}^0(Y^t, Y^*) - \textsc{BLEU}^0(Y^{t-1}, Y^*)$. Obviously, if $a_t=\textsc{read}$, no new words are written into $Y$, yielding $r_t^Q=0$. Note that we do not multiply $\textsc{BP}$
until the end of the sentence, as it would heavily penalize partial translation results.

\paragraph{Delay}As another critical feature, delay judges how much time is wasted waiting for the translation. Ideally we would directly measure the actual time delay incurred by waiting for the next word. For simplicity, however, we suppose it consumes the same amount of time listening  for one more word. We define two measurements, global and local, respectively:
\vspace{-2pt}
\begin{itemize}[leftmargin=*]
    \item \textbf{Average Proportion (AP)}: following the definition in \cite{cho2016can}, $X$, $Y$ are the source and decoded sequences respectively, and we use $s(\tau)$ to denote the number of source words been waited when decoding word $y_\tau$, 
    \vspace{-3pt}
    \begin{equation}
    \begin{split}
        &0 < d\left(X, Y\right) = \frac{1}{|X||Y|}\sum_{\tau}{s(\tau)}\leq 1 \\
        &d_t = 
        \left\{\begin{array}{ll}
        0    &     t < T\\
        d(X, Y)  &     t = T
        \end{array} \right. 
    \end{split}
    \vspace{-3pt}
    \end{equation}
    $d$ is a global delay metric, which defines the average waiting proportion of the source sentence when translating each word.\vspace{-5pt}
    \item \textbf{Consecutive Wait length (CW)}: in speech translation, listeners are also concerned with long silences during which no translation occurs. To capture this, we also consider on how many words were waited for (\textsc{read}) consecutively between translating two words. For each action, where we initially define $c_0=0$,
    \vspace{-3pt}
    \begin{equation}
        c_t = 
        \left\{\begin{array}{ll}
        c_{t-1} + 1    &     a_t = \textsc{read}\\
        0             &      a_t = \textsc{write}
        \end{array} \right. 
        \vspace{-5pt}
    \end{equation}
\item \textbf{Target Delay:}~
We further define ``target delay" for both $d$ and $c$ as $d^*$ and $c^*$, respectively, as different simultaneous translation applications may have different requirements on delay. In our implementation, the reward function for delay is written as:
\vspace{-3pt}
\begin{equation}
    \label{eq_rd}
    r_t^{D} = \alpha \cdot \left[\text{sgn}(c_t - c^*)+1\right] + \beta \cdot \lfloor d_t - d^* \rfloor_{+}
    \vspace{-3pt}
\end{equation}
where $\alpha \leq 0, \beta \leq 0$. 
\end{itemize}

\paragraph{Trade-off between quality and delay}
A good simultaneous translation system requires balancing the trade-off of translation quality and time delay. Obviously, achieving the best translation quality and the shortest translation delays are in a sense contradictory. In this paper, the trade-off is achieved by balancing the rewards $r_t = r_t^Q + r_t^D$ provided to the system, that is, by adjusting the coefficients $\alpha,\beta$ and the target delay $d^*, c^*$ in Eq.~\ref{eq_rd}. 

\subsection{Reinforcement Learning}

\paragraph{Policy Gradient}
We freeze the pre-trained parameters of an NMT model, and train the agent using the policy gradient~\cite{williams1992simple}. The policy gradient maximizes the following expected cumulative future rewards,
$     J = \mathbb{E}_{\pi_{\theta}}\left[\sum_{t=1}^T r_t\right], $ whose gradient is 
\begin{equation}
\label{eq.train}
   \nabla_{\theta}J=\mathbb{E}_{\pi_{\theta}}\left[\sum_{t'=1}^T\nabla_{\theta}\log \pi_{\theta}(a_{t'}|\cdot) R_t\right]
\end{equation}
$R_t=\sum_{k=t}^T\left[r_k^Q + r_k^D\right]$ is the cumulative future rewards for current observation and action. 
In practice, Eq.~\ref{eq.train} is estimated by sampling multiple action trajectories from the current policy $\pi_{\theta}$, collecting the corresponding rewards.

\paragraph{Variance Reduction}~
Directly using the policy gradient suffers from high variance, which makes learning unstable and inefficient. We thus employ the variance reduction techniques suggested by~\newcite{mnih2014neural}. We subtract from $R_t$ the output of a baseline network $b_{\varphi}$ to obtain $\hat{R}_t = R_t - b_{\varphi}\left(o_t\right)$, and centered re-scale the reward as $\tilde{R}_t = \frac{\hat{R}_t - b}{\sqrt{\sigma^2+\epsilon}}$ with a running average $b$ and standard deviation $\sigma$. The baseline network is trained to minimize the squared loss as follows:
\vspace{-3pt}
\begin{equation}
    L_{\varphi} = \mathbb{E}_{\pi_{\theta}}\left[ \sum_{t=1}^T\|R_t - b_{\varphi}\left(o_t\right)\|^2 \right]
    \vspace{-3pt}
\end{equation}

We also regularize the negative entropy of  the policy 
to facilitate exploration. 

The overall learning algorithm is summarized in Algorithm~\ref{algo2}. For efficiency, instead of updating with stochastic gradient descent~(SGD) on a single sentence, both the agent and the baseline are optimized using 
a minibatch of multiple sentences.

\begin{algorithm}[t]
\caption{Learning with Policy Gradient}
\label{algo2}
\begin{algorithmic}[1]
\Require{NMT system $\phi$, agent $\theta$, baseline $\varphi$}
\State Pretrain the NMT system $\phi$ using MLE;
\State Initialize the agent $\theta$;
\While{stopping criterion fails}
\State Obtain a translation pairs: $\{(X, Y^*)\}$;
\For{$(Y, S) \sim$ Simultaneous Decoding}
\For{$(o_t, a_t)$ in $S$}
\State Compute the quality: $r_t^Q$;
\State Compute the delay: $r_t^D$;
\State Compute the baseline: $b_{\varphi}\left(o_t\right)$;
\EndFor
\EndFor
\State Collect the future rewards: $\{R_t\}$;
\State Perform variance reduction: $\{\tilde{R}_t\}$;
\State Update: $\theta \gets \theta + \lambda_1 \nabla_{\theta}\left[J - \kappa\mathcal{H}(\pi_{\theta})\right]$
\State Update: $\varphi \gets \varphi - \lambda_2 \nabla_{\varphi}L$
\EndWhile
\end{algorithmic}
\end{algorithm}

\vspace{-3mm}
\section{Simultaneous Beam Search}
\label{sec:beamsearch}
\begin{figure}[t]
   	\centering
          	\vspace{-2pt}
          	\includegraphics[width=\linewidth]{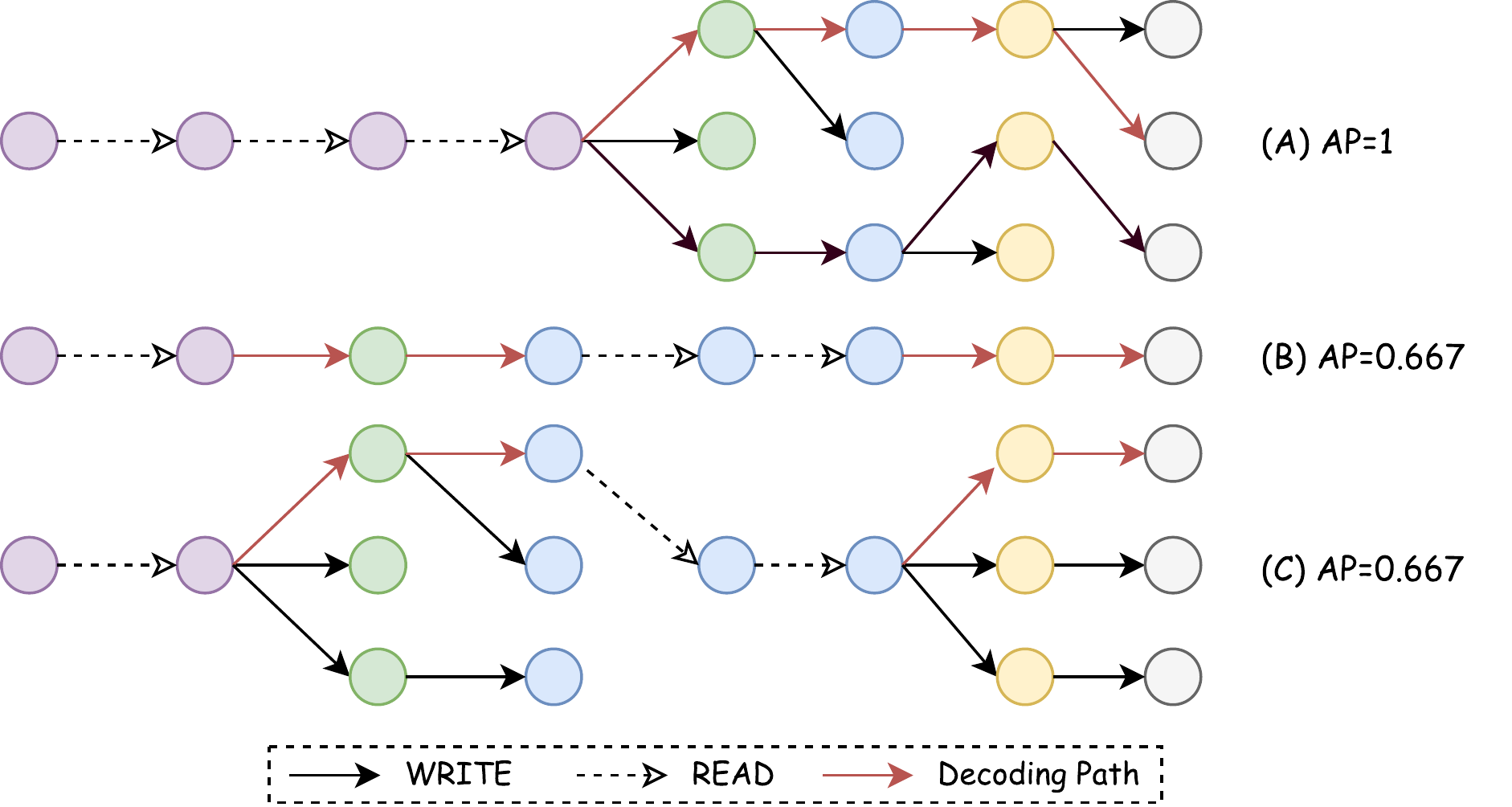} 
          	\caption{\label{beam} {Illustrations of (A)~beam-search, (B)~simultaneous greedy decoding and (C)~simultaneous beam-search.}} 
   \vspace{-4mm}
\end{figure}

\vspace{-1mm}
In previous sections we described a simultaneous greedy decoding algorithm. In standard NMT it has been shown that beam search, where the decoder keeps a beam of $k$ translation trajectories, greatly improves translation quality~\cite{sutskever2014sequence}, as shown in Fig.~\ref{beam}\,(A). 

It is non-trivial to directly apply beam-search in simultaneous machine translation,
as beam search waits until the last word to write down translation. Based on our assumption \textsc{write} does not cost delay,
we can perform a simultaneous beam-search when the agent chooses to consecutively \textsc{write}: keep multiple beams of translation trajectories in temporary buffer and output the best path when the agent switches to \textsc{read}. As shown in Fig.~\ref{beam} (B) \& (C), it tries to search for a relatively better path while keeping the delay unchanged.

Note that we do not re-train the agent for simultaneous beam-search. At each step we simply input the observation of the current best trajectory into the agent for making next decision. 

\section{Experiments}
\label{sec:experiments}
\begin{figure*}[!t]
\centering
\subfigure[BLEU (EN $\rightarrow$ RU)]{
\includegraphics[width=.31\textwidth]{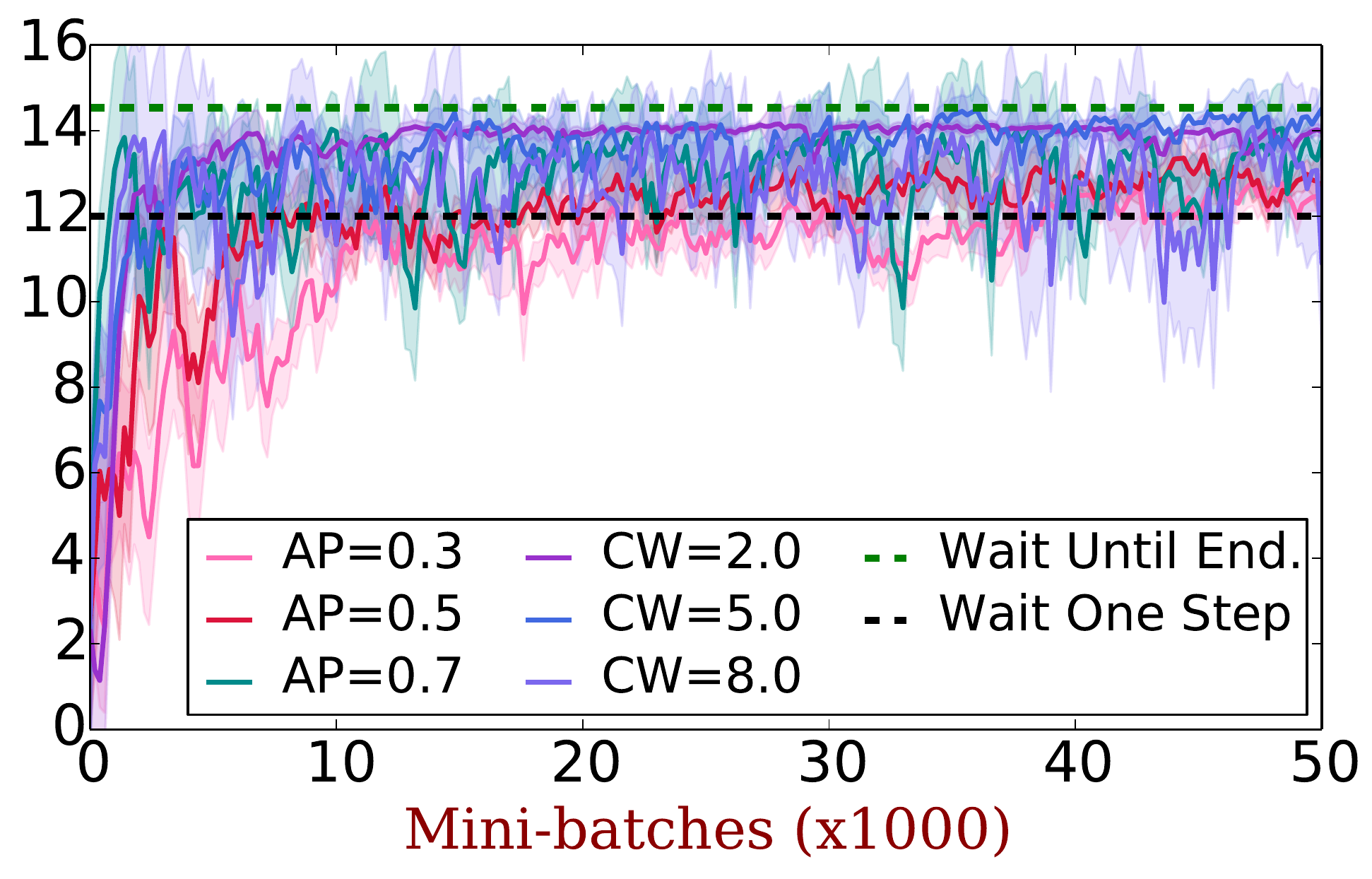}
}
\subfigure[AP (EN $\rightarrow$ RU)]{
\includegraphics[width=.31\textwidth]{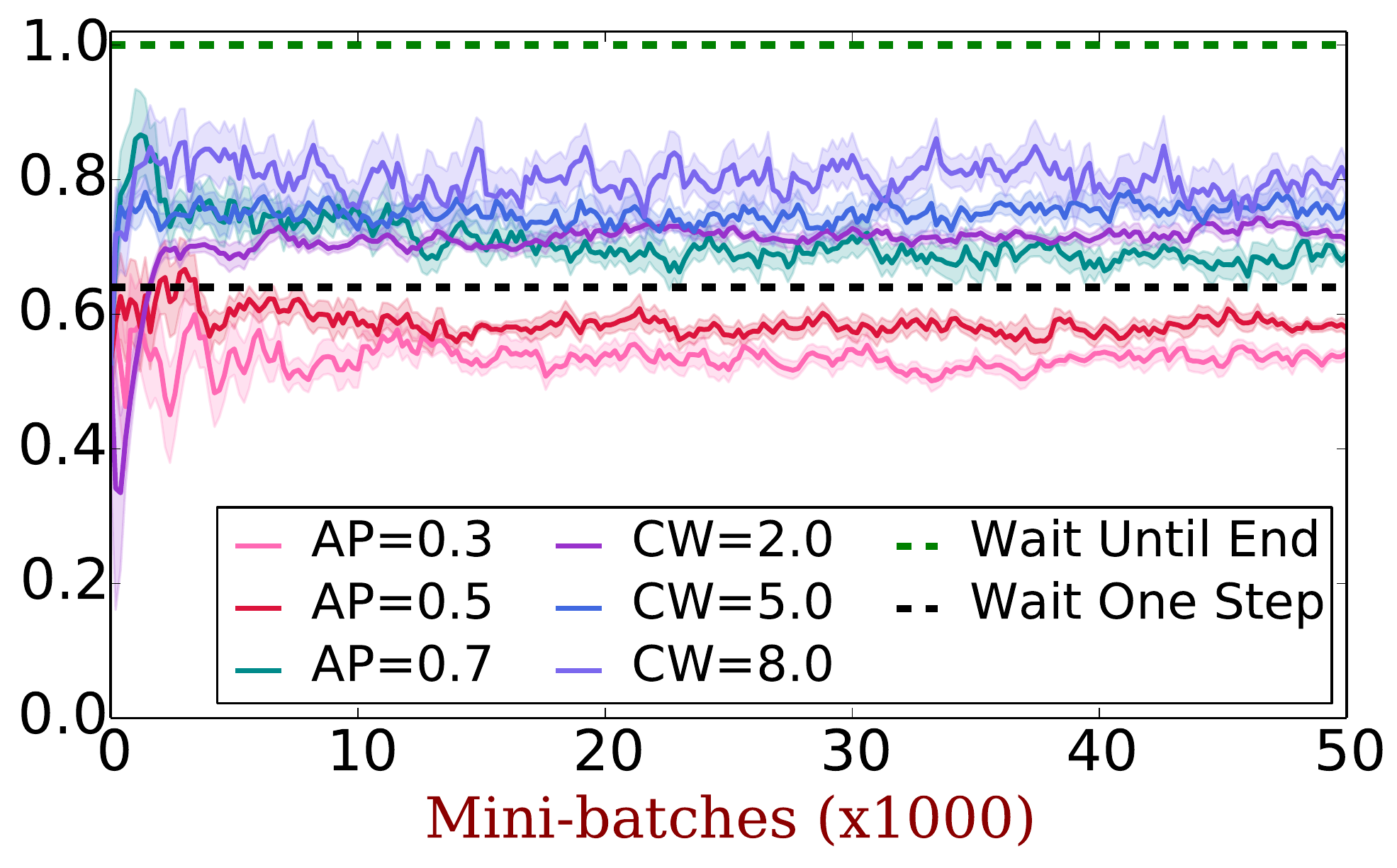}
}
\subfigure[CW (EN $\rightarrow$ RU)]{
\includegraphics[width=.31\textwidth]{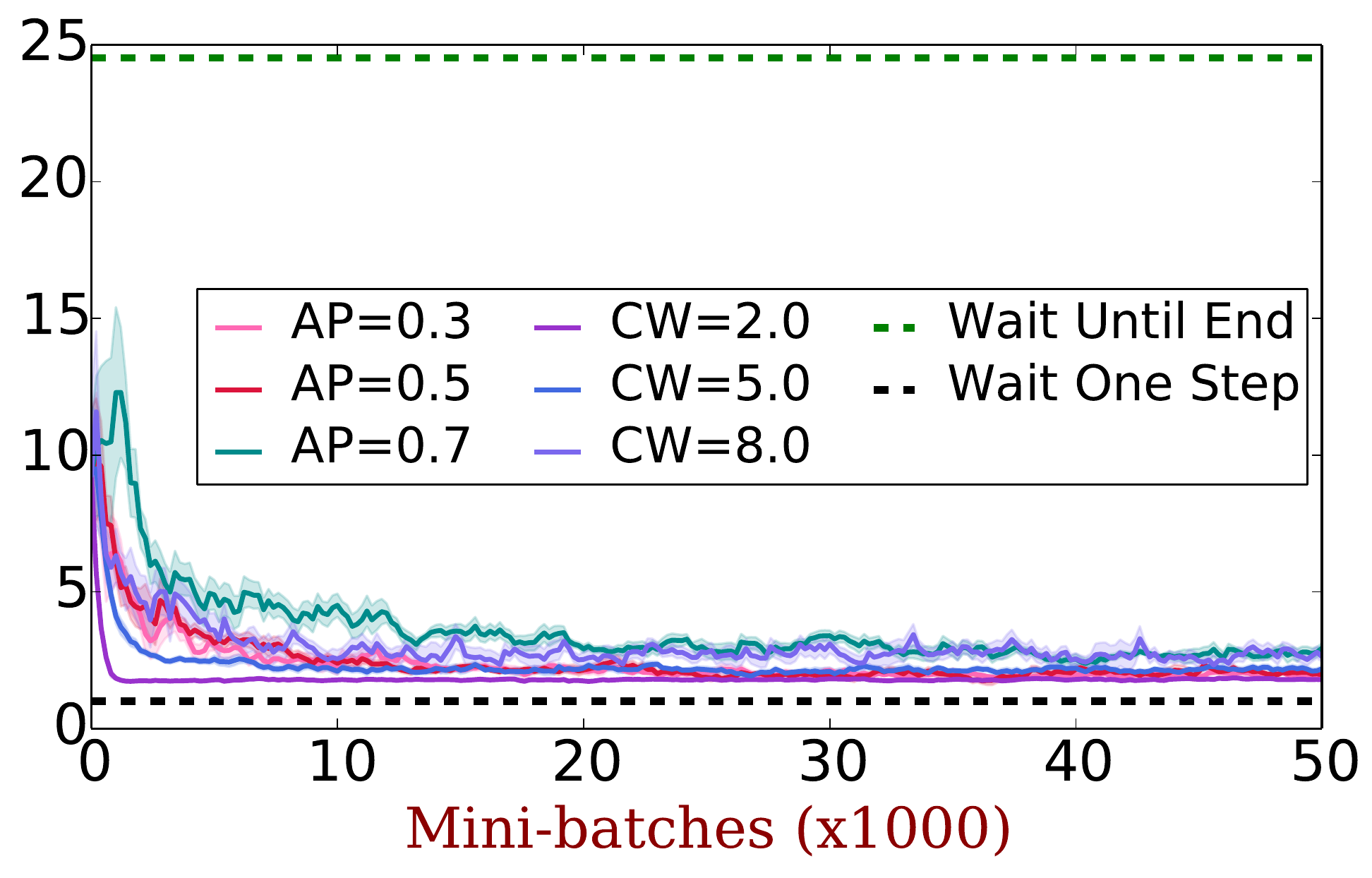}
}
\vspace{-4mm}
\caption{{Learning progress curves for variant delay targets on the validation dataset for EN $\rightarrow$ RU. Every time we only keep one target for one delay measure. For instance when using target AP, the coefficient of $\alpha$ in Eq.~\ref{eq_rd} will be set $0$.}}
\label{fig.lr}
\vspace{-7pt}
\end{figure*}

\begin{figure*}[htb]
\centering
\subfigure[EN$\rightarrow$RU]{
\includegraphics[width=.31\textwidth]{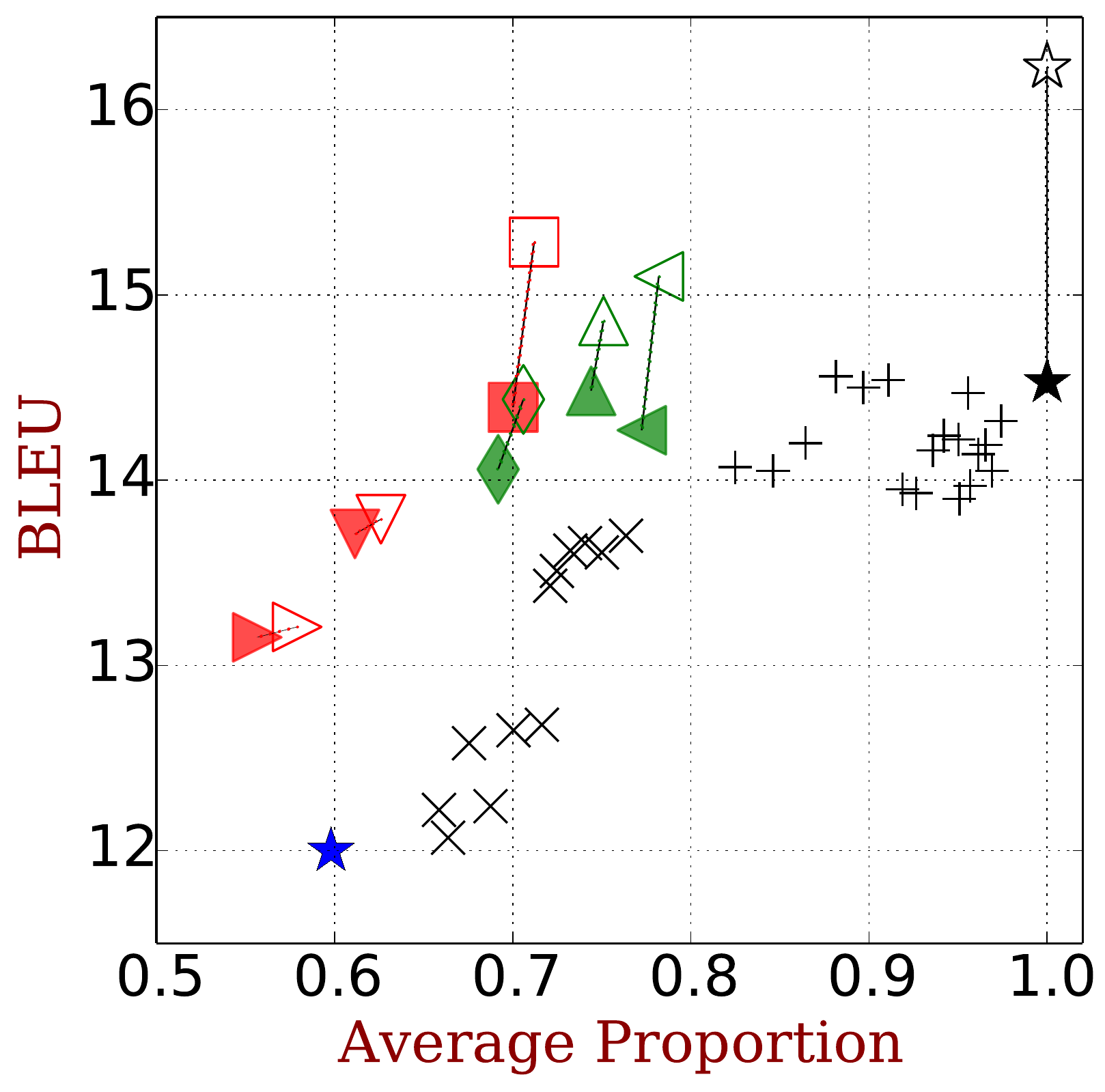}
}
\hfill
\subfigure[RU$\rightarrow$EN]{
\includegraphics[width=.31\textwidth]{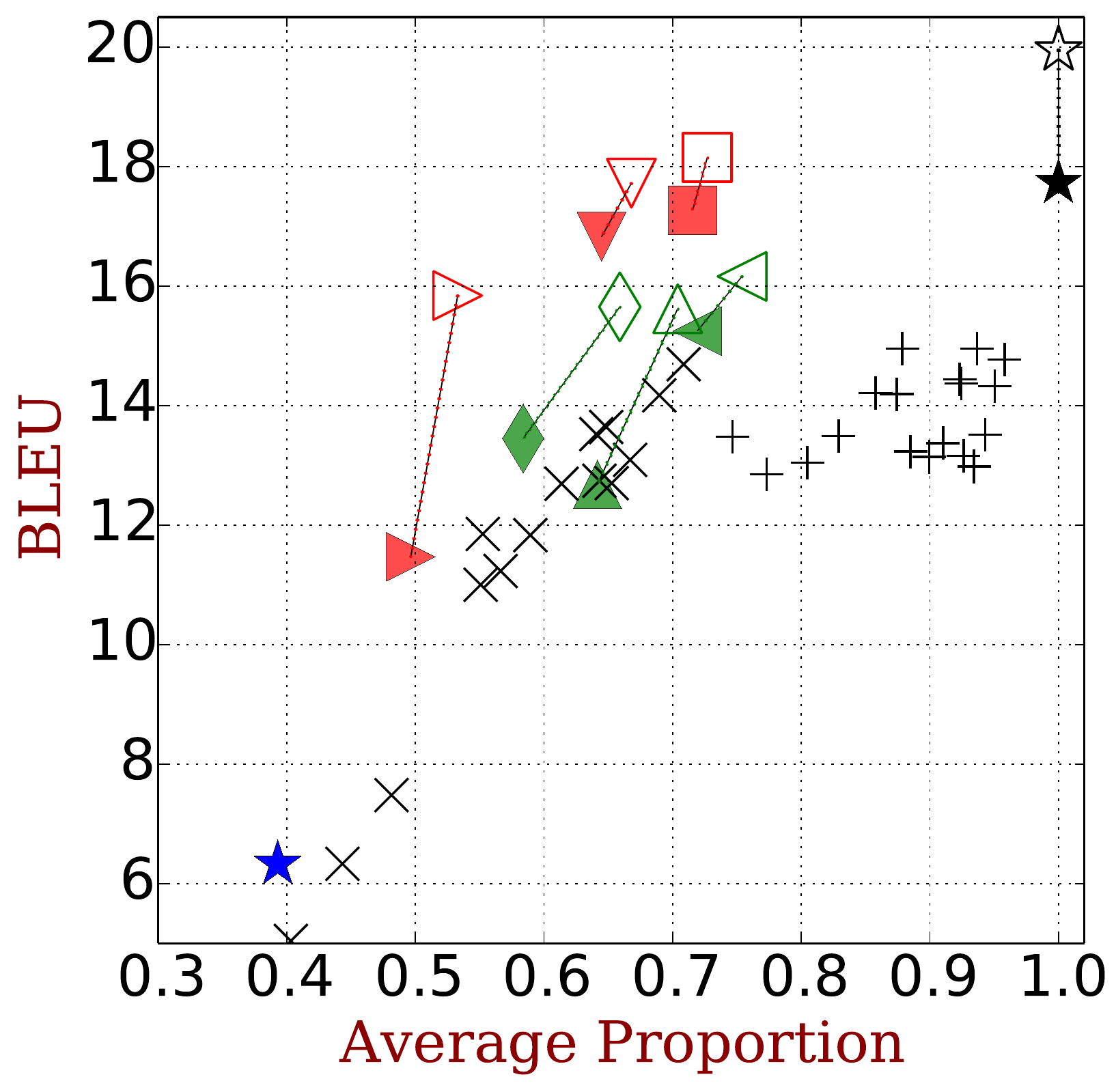}
}
\hfill
\subfigure[EN$\rightarrow$DE]{
\includegraphics[width=.31\textwidth]{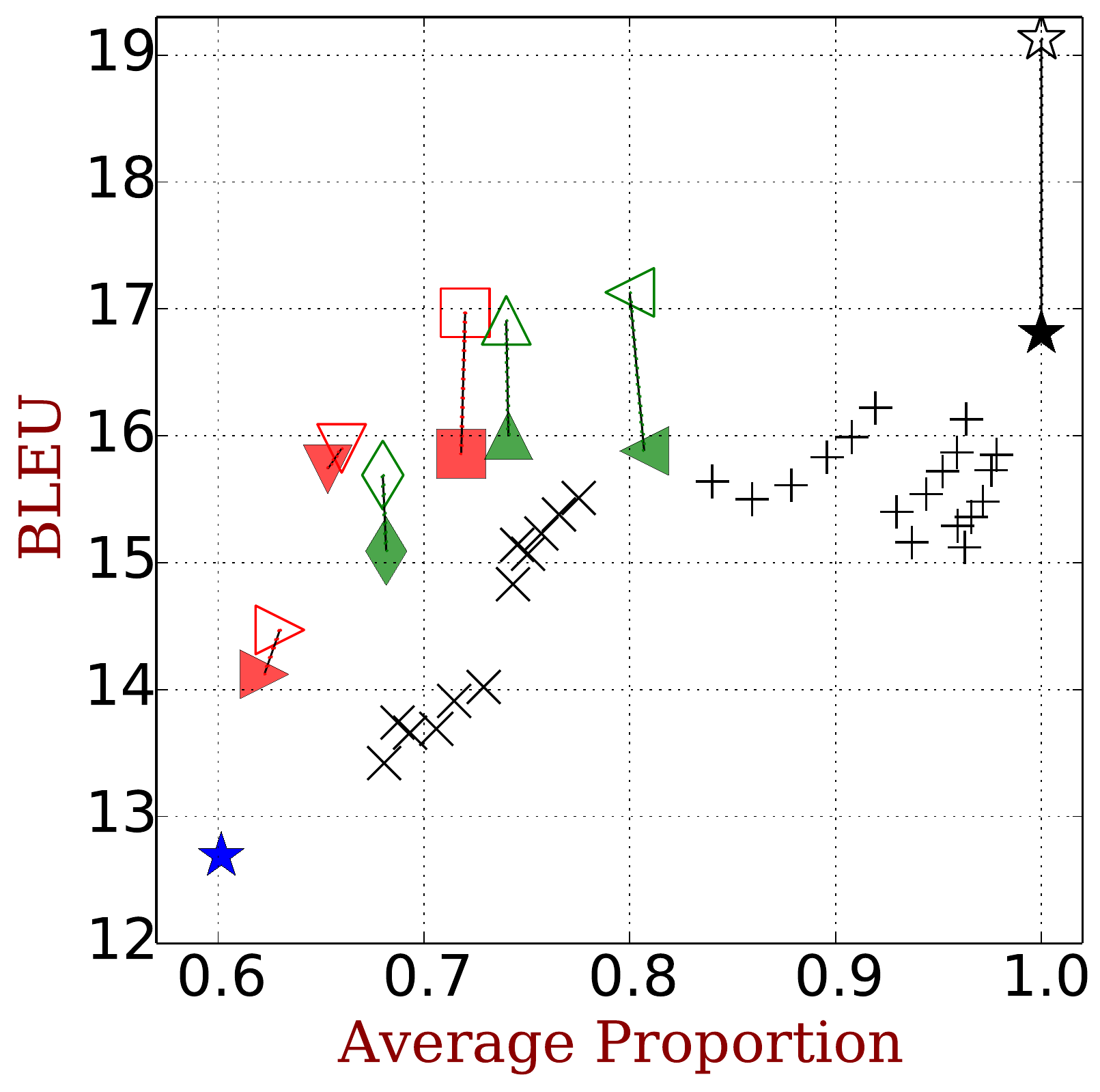}
}

\begin{minipage}{0.32\textwidth}
\subfigure[DE$\rightarrow$EN]{
\includegraphics[width=\textwidth]{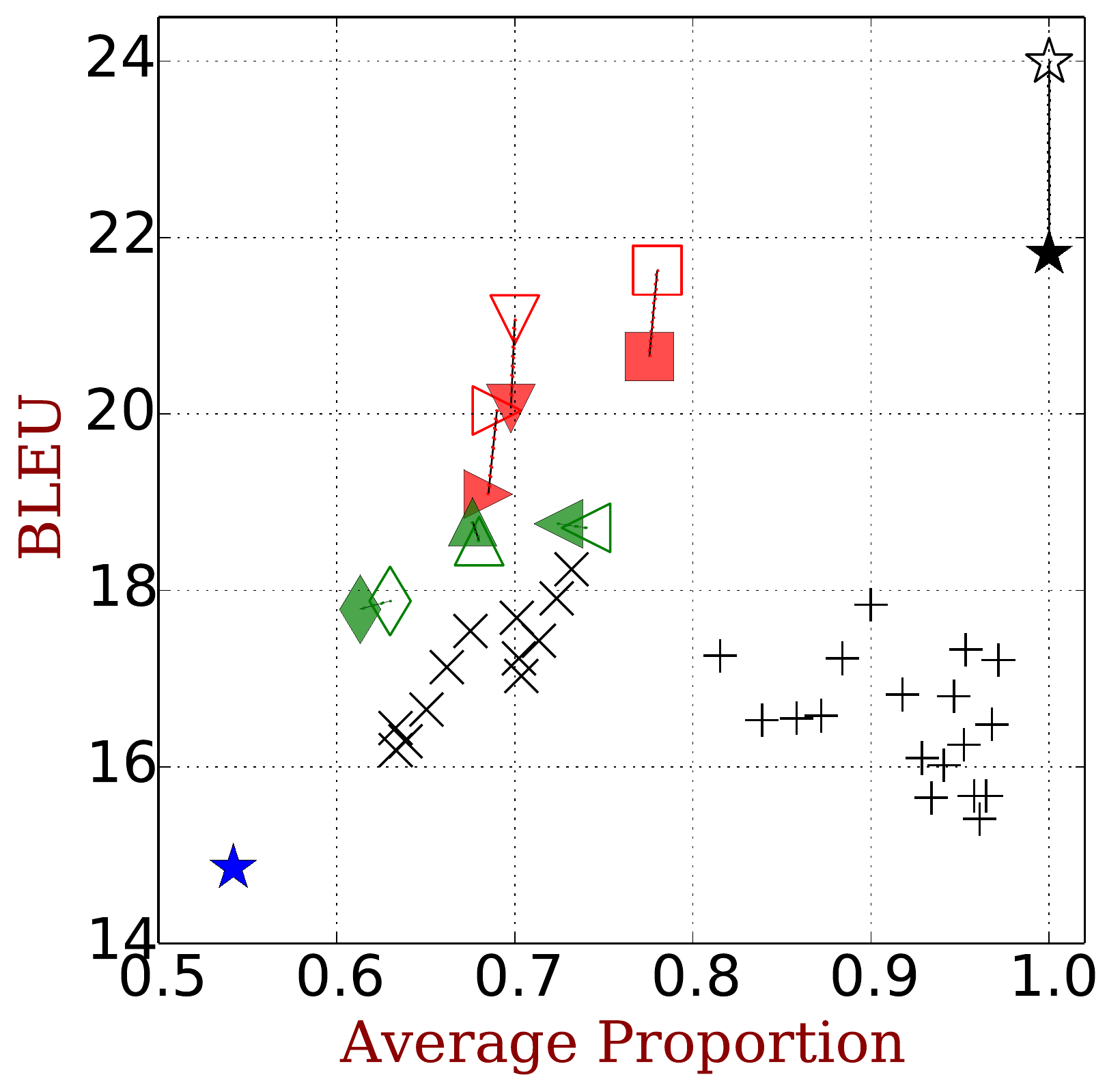}
}
\end{minipage}
\hfill
\begin{minipage}{0.58\textwidth}
\caption{{Delay~(AP) v.s. BLEU for both language pair--directions. The shown point-pairs are the results of simultaneous greedy decoding and beam-search (beam-size = 5) respectively with models trained for various delay targets: ($\color{green!70!blue}\blacktriangleleft\triangleleft$:   CW=$8$,
 $\color{green!70!blue}\blacktriangle\triangle$:           CW=$5$,
 $\color{green!70!blue}\blacklozenge\lozenge$:             CW=$2$,
 $\color{red}  \blacktriangleright\triangleright$: AP=$0.3$,
 $\color{red}  \blacktriangledown\triangledown$:   AP=$0.5$,
 $\color{red}  \blacksquare\square$:               AP=$0.7$)}. For each target, we select the model that maximizes the quality-to-delay ratio ($\frac{\text{BLEU}}{\text{AP}}$) on the validation set. The baselines are also plotted ($\color{blue}\bigstar$: WOS $\color{black}\bigstar$\ding{73}: WUE, $\times$: WID, $+$: WIW).}
\label{fig.bvd}
\end{minipage}
\hfill

\vspace{-6mm}
\end{figure*}

\subsection{Settings}
\paragraph{Dataset}~
To extensively study the proposed simultaneous translation model, we train and evaluate it on two different language pairs: ``English-German (EN-DE)" and ``English-Russian (EN-RU)" in both directions per pair. We use the parallel corpora available from WMT'15\footnote{http://www.statmt.org/wmt15/} for both pre-training the NMT environment and learning the policy. We utilize newstest-2013 as the validation set to evaluate the proposed algorithm. Both the training set and the validation set are tokenized and segmented into sub-word units with byte-pair encoding (BPE)~\cite{sennrich2015neural}. We only use sentence pairs where both sides are less than 50 BPE subword symbols long for training.
\vspace{-3pt}
\paragraph{Environment \& Agent Settings}
We pre-trained the NMT environments for both language pairs and both directions following the same setting from~\cite{cho2016can}. 
We further built our agents, using a recurrent policy with 512 GRUs and a softmax function to produce the action distribution. All our agents are trained using policy gradient using Adam~\cite{kingma2014adam} optimizer, with a mini-batch size of 10. For each sentence pair in a batch, 5 trajectories are sampled. 
For testing, instead of sampling we pick the action with higher probability each step.


\vspace{-3pt}
\paragraph{Baselines}~ We compare the proposed methods against previously proposed baselines.
For fair comparison, we use the same NMT environment:
\begin{itemize}[leftmargin=*]
\vspace{-2pt}
\item \textbf{Wait-Until-End (WUE)}: an agent that starts to \textsc{write} only when the last source word is seen. In general, we expect this to achieve the best quality of translation. We perform both greedy decoding and beam-search with this method.
\vspace{-2pt}
\item \textbf{Wait-One-Step (WOS)}: an agent that \textsc{write}s after each \textsc{read}s. 
Such a policy is problematic when the source and target language pairs have different word orders or lengths (e.g. EN-DE). 
\vspace{-2pt}
\item \textbf{Wait-If-Worse/Wait-If-Diff (WIW/WID)}: as proposed by \newcite{cho2016can}, the algorithm first pre-\textsc{read}s the next source word, and accepts this \textsc{read} when the probability of the most likely target word decreases~(WIW), or the most likely target word changes~(WID). 
\vspace{-2pt}
\item \textbf{Segmentation-based (SEG)}~\cite{oda-EtAl:2014:P14-2}: a state-of-the-art segmentation-based algorithm based on optimizing segmentation to achieve the highest quality score. In this paper, we tried the simple greedy method~(SEG1) and the greedy method with POS Constraint~(SEG2).
\end{itemize}

\subsection{Quantitative Analysis}

In order to evaluate the effectiveness of our reinforcement learning algorithms with different reward functions, we vary the target delay $d^* \in \{0.3, 0.5, 0.7\}$ and $c^* \in \{2, 5, 8\}$ for Eq.~\ref{eq_rd} separately, and trained agents with $\alpha$ and $\beta$ adjusted to values that provided stable learning for each language pair according to the validation set.
\vspace{-3pt}
\paragraph{Learning Curves}~ As shown in Fig.~\ref{fig.lr}, we plot learning progress for EN-RU translation with different target settings. It clearly shows that our algorithm effectively increases translation quality for all the models, while pushing the delay close, if not all of the way, to the target value.
It can also be noted from Fig.~\ref{fig.lr} (a) and (b) that there exists strong correlation between the two delay measures, implying the agent can learn to decrease both AP and CW simultaneously.

\paragraph{Quality v.s. Delay}~ As shown in Fig.~\ref{fig.bvd}, it is clear that the trade-off between translation quality and delay has very similar behaviors across both language pairs and directions. The smaller delay (AP or CW) the learning algorithm is targeting, the lower quality (BLEU score) the output translation. It is also interesting to observe that, it is more difficult for ``$\rightarrow$EN'' translation to achieve a lower AP target while maintaining good quality, compared to ``EN$\rightarrow$''. 
In addition, the models that are optimized on AP tend to perform better than those optimized on CW, especially in ``$\rightarrow$EN'' translation. German and Russian sentences tend to be longer than English, hence require more consecutive waits before being able to emit the next English symbol.

\begin{figure}[t]
   	\centering
          	\includegraphics[width=1.0\linewidth]{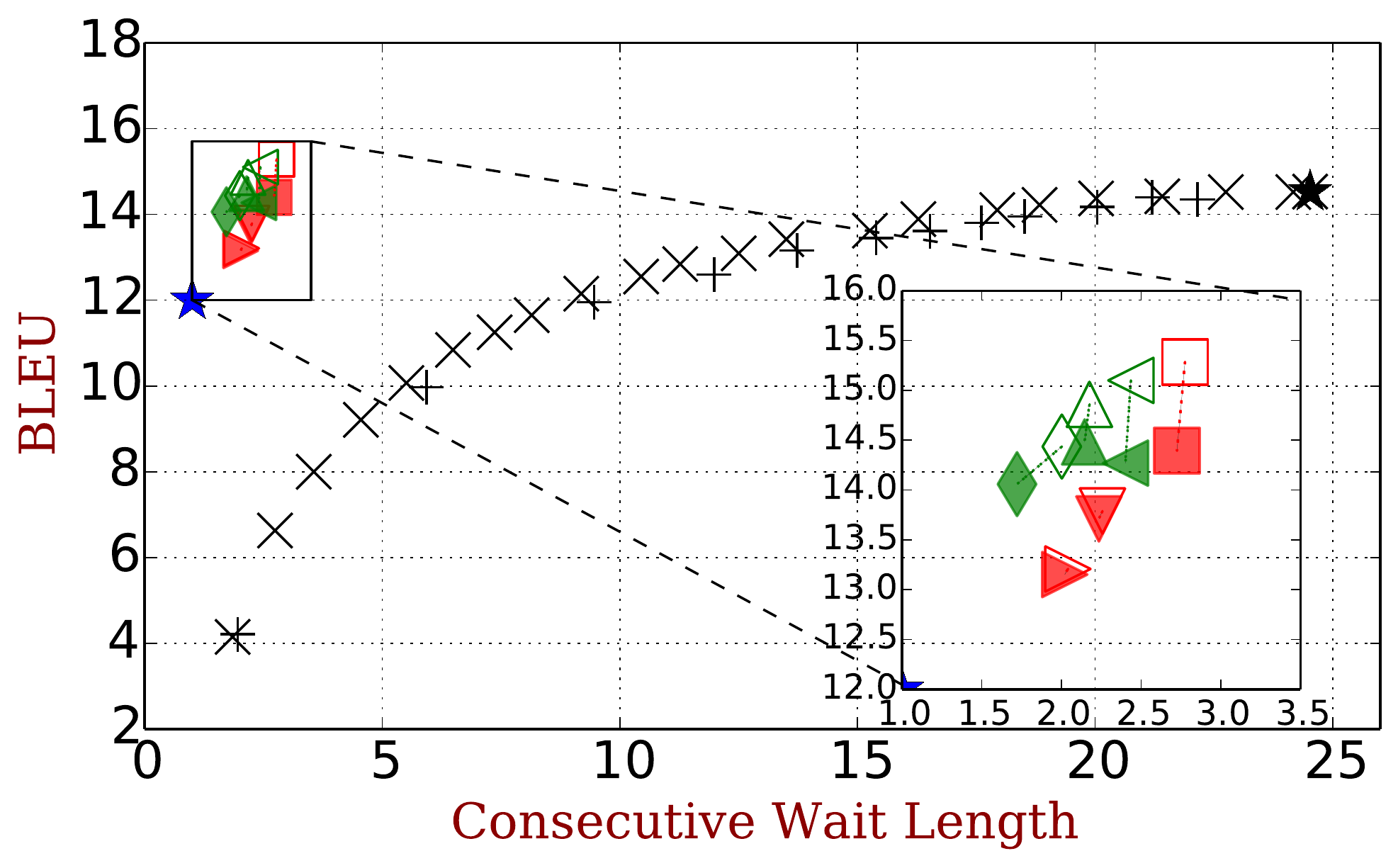} 
          	\vspace{-20pt}
          	\caption{\label{wait} {Delay~(CW) v.s. BLEU score for EN $\rightarrow$ RU,
($\color{green!70!blue}\blacktriangleleft\triangleleft$:   CW=$8$,
 $\color{green!70!blue}\blacktriangle\triangle$:           CW=$5$,
 $\color{green!70!blue}\blacklozenge\lozenge$:             CW=$2$,
 $\color{red}  \blacktriangleright\triangleright$: AP=$0.3$,
 $\color{red}  \blacktriangledown\triangledown$:   AP=$0.5$,
 $\color{red}  \blacksquare\square$:               AP=$0.7$), against the baselines~($\color{blue}\bigstar$: WOS $\color{black}\bigstar$: WUE, $+$: SEG1, $\times$: SEG2).}} 
 \vspace{-10pt}
\end{figure}

\vspace{-3pt}
\paragraph{v.s. Baselines}~
In Fig.~\ref{fig.bvd} and~\ref{wait}, 
the points closer to the upper left corner achieve better trade-off performance. 
Compared to WUE and WOS which can ideally achieve the best quality (but the worst delay) and the best delay (but poor quality) respectively, all of our proposed models find a good balance between quality and delay.  Some of the proposed models can achieve good BLEU scores close to WUE, while have much smaller delay.

Compared to the method of \newcite{cho2016can} based on two hand-crafted rules (WID, WIW), in most cases our proposed models find better trade-off points, while there are a few exceptions. We also observe that the baseline models have trouble controlling the delay in a reasonable area. In contrast, by optimizing towards a given target delay, our proposed model is stable while maintaining good translation quality.
  
We also compared against \newcite{oda-EtAl:2014:P14-2}'s state-of-the-art segmentation algorithm (SEG). As shown in Fig~\ref{wait}, it is clear that although SEG can work with variant segmentation lengths (CW), 
the proposed model outputs high quality translations at a much smaller CW. 
We conjecture that this is due to the independence assumption in SEG,
while the RNNs and attention mechanism in our model makes it possible to look at the whole history to decide each translated word.
\vspace{-6pt}
\paragraph{w/o Beam-Search}
We also plot the results of simultaneous beam-search instead of using greedy decoding. 
It is clear from Fig.~\ref{fig.bvd} and \ref{wait} that most of the proposed models can achieve an visible increase in quality together with a slight increase in delay. 
This is because beam-search can help to avoid bad local minima.  
We also observe that the simultaneous beam-search cannot bring as much improvement as it did in the standard NMT setting. In most cases, the smaller delay the model achieves, the less beam search can help as it requires longer consecutive \textsc{write} segments for extensive search to be necessary.
One possible solution is to consider the beam uncertainty in the agent's \textsc{read/write} decisions. We leave this to future work. 

\subsection{Qualitative Analysis}
\begin{figure*}[!t]
\centering
\subfigure[Simultaneous Neural Machine Translation]{
\includegraphics[width=0.48\textwidth]{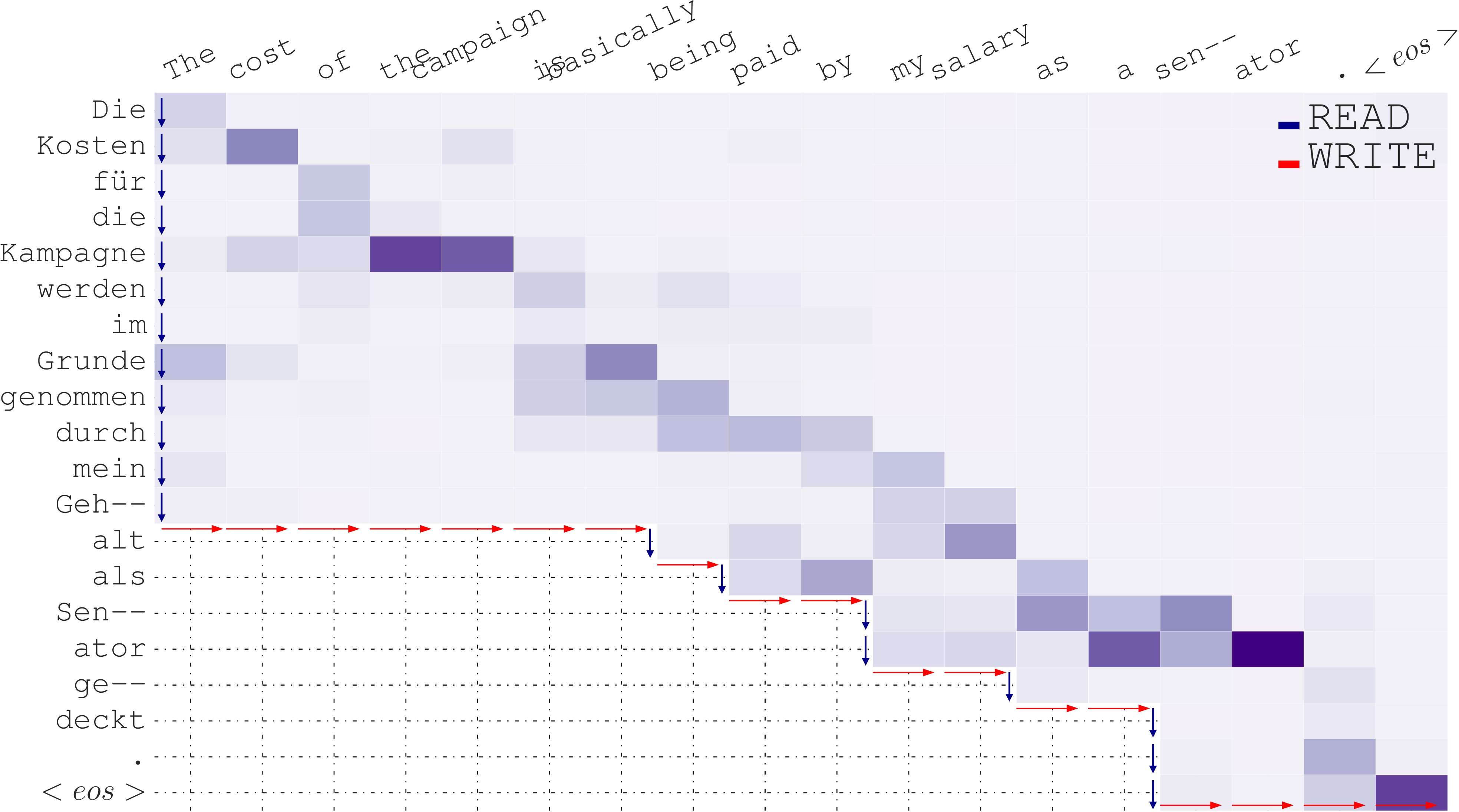}
}
\subfigure[Neural Machine Translation]{
\includegraphics[width=0.48\textwidth]{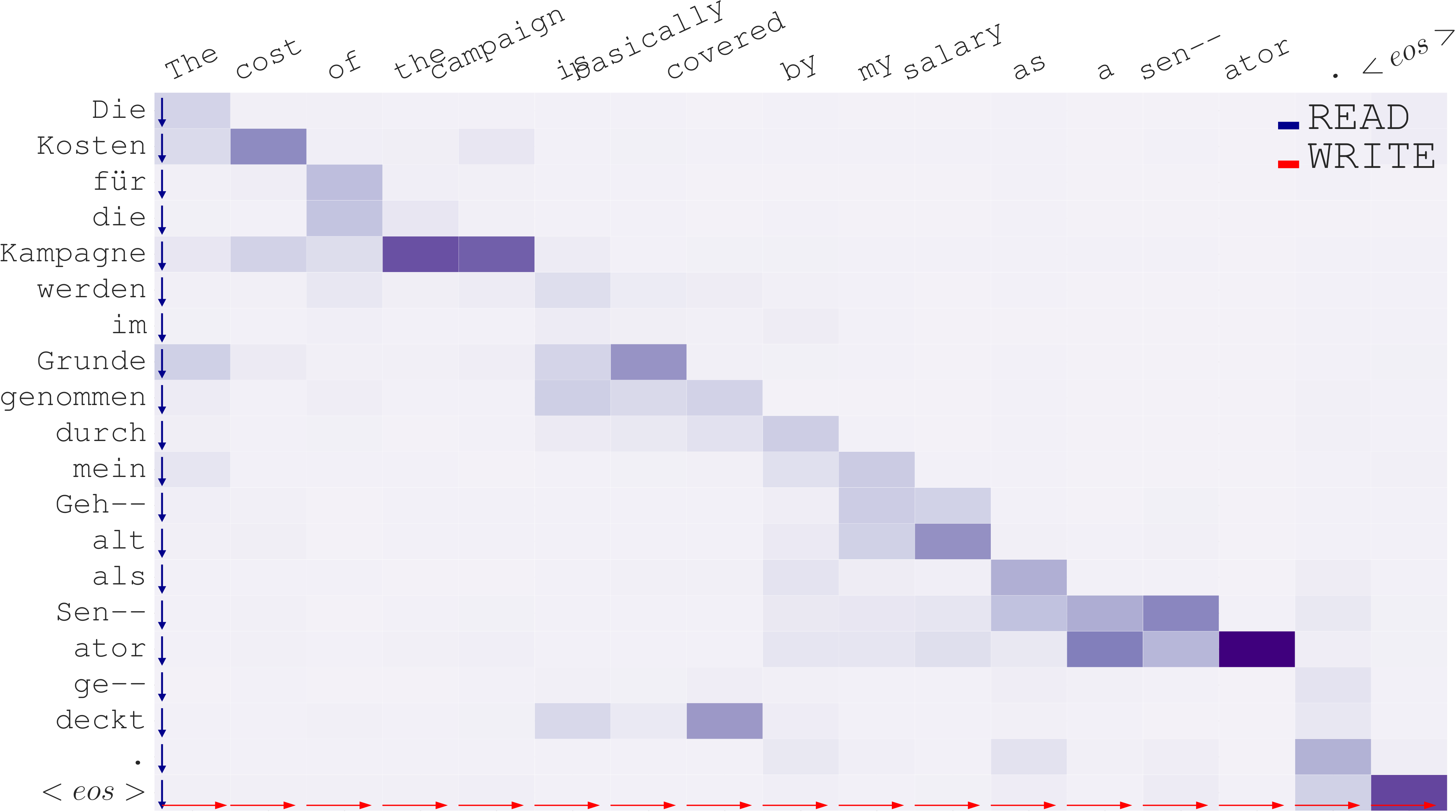}
}
\caption{\label{deen2}{Comparison of DE$\rightarrow$EN examples using the proposed framework and usual NMT system respectively. Both the heatmaps share the same setting with Fig.~\ref{crop}}. The verb ``gedeckt'' is incorrectly translated in simultaneous translation.}
\end{figure*}

\begin{figure*}[!t]
   	\centering
  	\includegraphics[width=\linewidth]{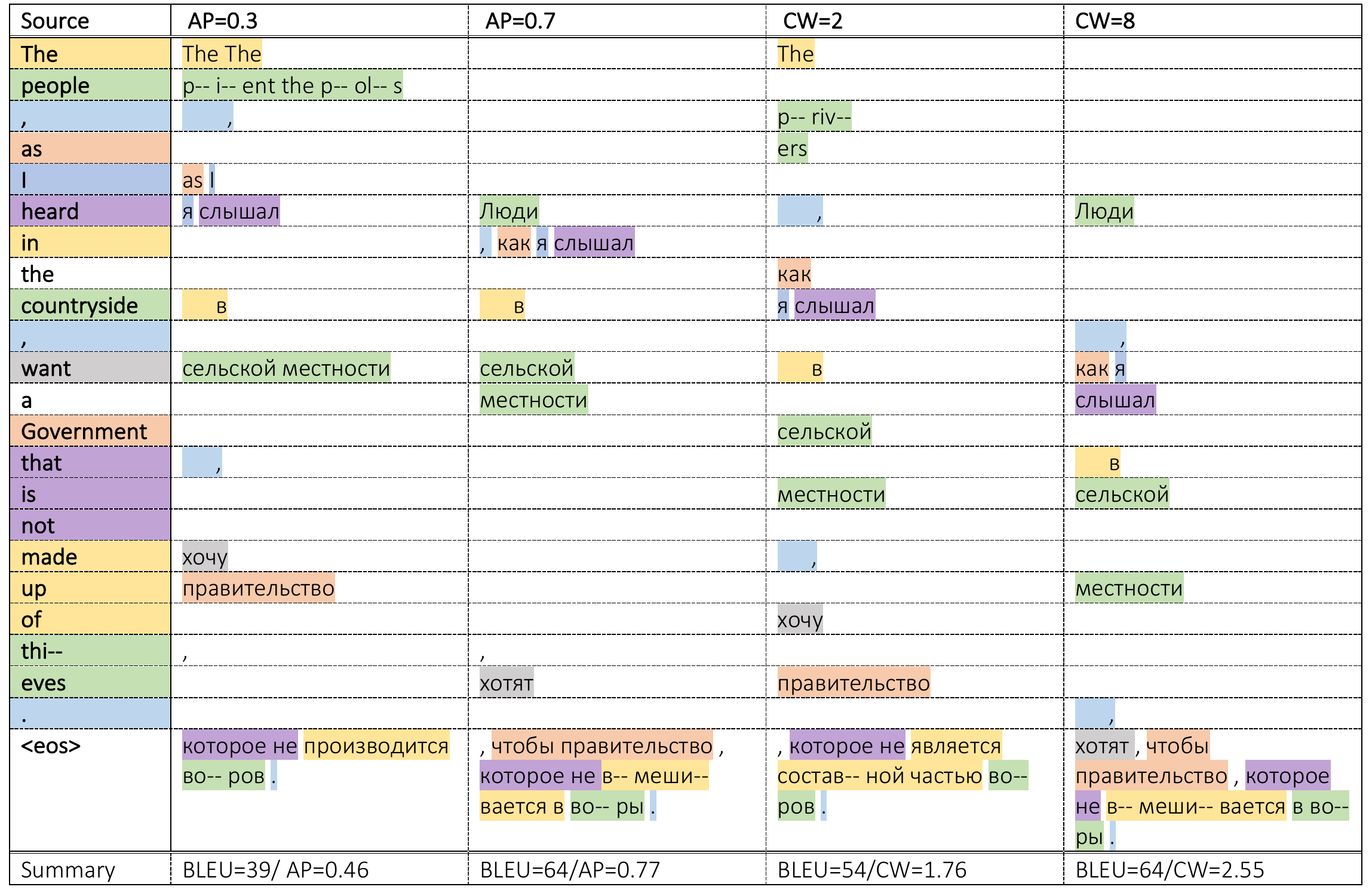} 
          	\caption{\label{exp1} {Given the example input sentence (leftmost column), we show outputs by models trained for various delay targets. For these outputs, each row corresponds to one source word and represents the emitted words (maybe empty) after reading this word. The corresponding source and target words are in the same color for all model outputs.}}
\end{figure*} 

In this section, we perform a more in-depth analysis using examples from both EN-RU and EN-DE pairs, in order to have a deeper understanding of the proposed algorithm and its remaining limitations. We only perform greedy decoding to simplify visualization. 
\vspace{-5pt}
\paragraph{EN$\rightarrow$RU}~ As shown in Fig~\ref{exp1}, since both English and Russian are Subject-–Verb-–Object~(SVO) languages, the corresponding words may share the same order in both languages, which makes simultaneous translation easier. 
It is clear that the larger the target delay (AP or CW) is set, the more words are read before translating the corresponding words, which in turn results in better translation quality. 
We also note that very early \textsc{write} commonly causes bad translation. For example, for AP=0.3 \& CW=2, both the models choose to \textsc{write} in the very beginning the word ``The'', which is unreasonable since Russian has no articles, and there is no word corresponding to it. One good feature of using NMT is that the more words the decoder \textsc{read}s, the longer history is saved, rendering simultaneous translation easier.


\vspace{-5pt}
\paragraph{DE$\rightarrow$EN} 
As shown in Fig~\ref{crop} and~\ref{deen2}~(a), where we visualize the attention weights as soft alignment between the progressive input and output sentences, the highest weights are basically along the diagonal line. This indicates that our simultaneous translator works by waiting for enough source words with high alignment weights and then switching to write them. 

DE-EN translation is likely more difficult as German usually uses Subject-Object-Verb~(SOV) constructions a lot. 
As shown in Fig~\ref{crop},
when a sentence (or a clause) starts the agent has learned such policy to \textsc{read} multiple steps to approach the verb (e.g. serviert and gestorben in Fig~\ref{crop}). Such a policy is still limited when the verb is very far from the subject. For instance in Fig.~\ref{deen2}, the simultaneous translator achieves almost the same translation with standard NMT except for the verb ``gedeckt'' which corresponds to ``covered'' in NMT output. Since there are too many words between the verb ``gedeckt'' and the subject ``Kosten f\"{u}r die Kampagne werden'', the agent gives up reading (otherwise it will cause a large delay and a penalty) and \textsc{write}s ``being paid'' based on the decoder's hypothesis. This is one of the limitations of the proposed framework, as the NMT environment is trained on complete source sentences and it may be difficult to predict the verb that has not been seen in the source sentence. 
One possible way is to fine-tune the NMT model on incomplete sentences to boost its prediction ability. We will leave this as future work.

\section{Related Work}
Researchers commonly consider the problem of simultaneous machine translation in the scenario of real-time speech interpretation~\cite{fugen2007simultaneous,bangalore2012real,fujita2013simple,sridhar2013segmentation,yarmohammadi2013incremental}. In  this approach, the incoming speech stream required to be translated are first recognized and segmented based on an automatic speech recognition (ASR) system. The translation model then works independently based on each of these segments, potentially limiting the quality of translation. 
To avoid using a fixed segmentation algorithm, \newcite{oda-EtAl:2014:P14-2} introduced a trainable segmentation component into their system, so that the segmentation leads to better translation quality. \newcite{grissomii2014don} proposed a similar framework, however, based on reinforcement learning. All these methods still rely on translating each segment independently without previous context.

Recently, two research groups have tried to apply the NMT framework to the simultaneous translation task. \newcite{cho2016can} proposed a similar waiting process. However, their waiting criterion is manually defined without learning. \newcite{satija2016simultaneous} proposed a method similar to ours in overall concept, but it significantly differs from our proposed method in many details. The biggest difference is  that they proposed to use an agent that passively reads a new word at each step. Because of this, it cannot consecutively decode multiple steps, rendering beam search difficult. In addition, they lack the comparison to any existing approaches. On the other hand, we perform an extensive experimental evaluation against state-of-the-art baselines, demonstrating the relative utility both quantitatively and qualitatively.

The proposed framework is also related to some recent efforts about online sequence-to-sequence (\textsc{seq2seq}) learning. 
\newcite{jaitly2015online} proposed a \textsc{seq2seq} ASR model that takes fixed-sized segments of the input sequence and outputs tokens based on each segment in real-time. It is trained with alignment information using supervised learning. A similar idea for online ASR is proposed by \newcite{luo2016learning}. Similar to \newcite{satija2016simultaneous}, they also used reinforcement learning to decide whether to emit a token while reading a new input at each step. Although sharing some similarities, ASR is very different from simultaneous MT with a more intuitive definition for segmentation. 
In addition, \newcite{yu2016online} recently proposed an online alignment model to help sentence compression and morphological inflection. They regarded the alignment between the input and output sequences as a hidden variable, and performed transitions over the input and output sequence. By contrast, the proposed \textsc{read} and \textsc{write} actions do not necessarily to be performed on aligned words (e.g. in Fig.~\ref{crop}), and are learned to balance the trade-off of quality and delay.

\section{Conclusion}
We propose a unified framework to do neural simultaneous machine translation. 
To trade off quality and delay, we extensively explore various targets for delay and design a method for beam-search applicable in the simultaneous MT setting.
Experiments against state-of-the-art baselines on two language pairs demonstrate the efficacy both quantitatively and qualitatively.

\section*{Acknowledgments}
KC acknowledges the support by Facebook, Google (Google Faculty Award 2016) and NVidia (GPU Center of Excellence 2015-2016). GN acknowledges the support of the Microsoft CORE program. This work was also partly supported by  Samsung Electronics (Project: "Development and Application of Larger-Context Neural Machine Translation").



\bibliography{eacl2017}
\bibliographystyle{eacl2017}







\end{document}